\documentclass[10pt, conference]{IEEEtran}

\IEEEoverridecommandlockouts                              

\usepackage{amsmath} 
\usepackage{amssymb}
\usepackage{booktabs}
\usepackage{multirow}
\usepackage{array}
\usepackage{textcomp}   
\usepackage{balance}
\usepackage[caption=false,font=footnotesize]{subfig}
\usepackage{booktabs}
\usepackage{xcolor}
\usepackage{subcaption}
\usepackage{graphicx}
\usepackage{subcaption}
\usepackage{hyperref}
\usepackage[dvipsnames,table,xcdraw]{xcolor}
\usepackage{pifont}
\usepackage[numbers,sort&compress]{natbib}

\usepackage{tabularx, boldline}
\usepackage{cellspace}
\usepackage{makecell}
\usepackage{soul}
\usepackage[symbol]{footmisc}

\title{\LARGE \bf
Massive Parallel Deep Reinforcement Learning for Active SLAM
}

\author{Martín Arce Llobera\textsuperscript{1}, Julio A. Placed\textsuperscript{2}, Mariano De Paula\textsuperscript{3}, and Pablo De Cristóforis\textsuperscript{1}
\thanks{This work was partially supported by DGA\_FSE T73\_23R, and UBACyT (20020220200126BA).}
\thanks{$^{1}$Martín Arce Llobera and Pablo De Cristóforis are with Facultad de Ciencias Exactas y Naturales, Universidad de Buenos Aires, Ciudad Autonoma de Buenos Aires, C1428EGA, Argentina ({\tt\footnotesize \{marcellobera,pdecris\}@dc.uba.ar}).}
\thanks{$^{2}$Julio A.~Placed is with the Instituto Tecnol\'ogico de Arag\'on (ITA) and the University of Zaragoza, Mar\'ia de Luna 3-7, Zaragoza, Spain ({\tt\footnotesize jplaced@ita.es}).}
\thanks{$^{3}$Mariano De Paula is with  INTELYMEC, Centro de Investigaciones en Física e Ingeniería del Centro (CIFICEN), UNICEN-CICPBA-CONICET, Olavarría, Buenos Aires, Argentina ({\tt\footnotesize mariano.depaula@fio.unicen.edu.ar}).} 
}

\begin{document}

\maketitle
\thispagestyle{empty}
\pagestyle{empty}

\begin{abstract}


Recent advances in parallel computing and GPU acceleration have created new opportunities for computation-intensive learning problems such as Active SLAM ---where actions are selected to reduce uncertainty and improve joint mapping and localization. However, existing DRL-based approaches remain constrained by the lack of scalable parallel training.


In this work, we address this challenge by proposing a scalable end-to-end DRL framework for Active SLAM that enables massively parallel training.
Compared with the state of the art, our method significantly reduces training time, supports continuous action spaces and facilitates the exploration of more realistic scenarios. It is released as an open-source framework\footnote[2]{\texttt{https://anonymous.4open.science/r/pdrl-aslam-8BE7}} to promote reproducibility and community adoption.

\end{abstract}

\section{Introduction}

Autonomous exploration of unknown environments can be formulated as a decision-making problem in which a robot selects actions to maximize its knowledge of the environment.  
When no external localization system is available, the robot must simultaneously estimate its pose while building a map of the surroundings, a problem known as Simultaneous Localization and Mapping (SLAM)~\cite{cadena2016past}. As the robot moves, however, uncertainty accumulates in its pose estimate, which in turn degrades the accuracy of the constructed map.
Actions are therefore needed to counteract this problem, such as navigating to highly informative areas and revisiting previously traversed regions (loop closure) to refine both robot’s localization and the map. Nevertheless, in SLAM, the robot’s trajectory is assumed to be predefined, or the robot is controlled by a human operator.
Thus, the effect of the robot choosing different actions on the accuracy of its location and the quality of the constructed map is not explicitly considered.

Active SLAM (ASLAM)~\cite{feder1999adaptive} extends the SLAM problem by incorporating decision-making into the estimation process. It can be defined as controlling a robot performing SLAM in order to reduce uncertainty in both its pose and map representation while exploring an unknown environment. Typically, ASLAM involves identifying candidate exploration locations, evaluating the expected information gain of possible actions, and selecting the action that maximizes a utility measure~\cite{placed2023survey}.

\begin{figure}[!h]
\centering
\includegraphics[width=.85\linewidth]{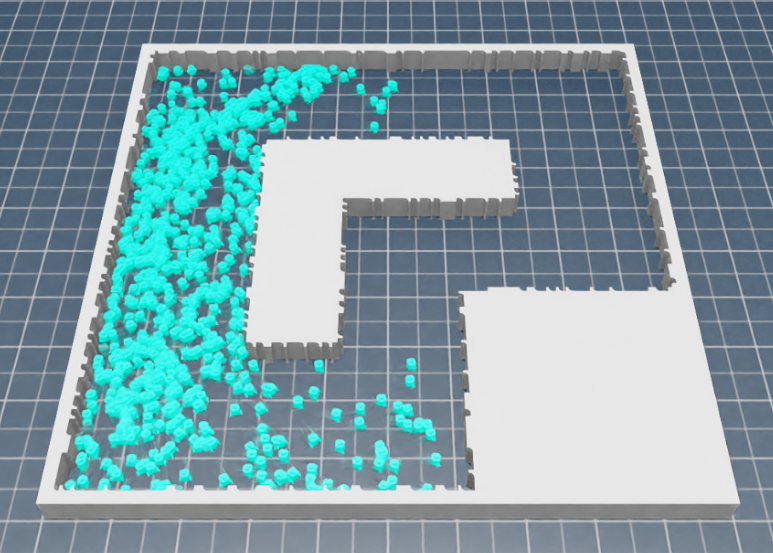}
\caption{
Parallel DRL training with 750 robots (cyan). The proposed pipeline leverages NVIDIA Isaac Sim/Lab for Active SLAM training in a GPU-accelerated environment.}
\label{framework}
\end{figure}

Deep Reinforcement Learning (DRL) has emerged as a promising approach for ASLAM due to its ability to handle partial observability and high-dimensional sensory inputs. Deep neural networks can process noisy observations such as LiDAR or camera data while implicitly representing the agent's \textit{belief} about the true state. Moreover, DRL enables learning exploration strategies directly from interaction with the environment, without requiring explicit models of dynamics or observation processes~\cite{arulkumaran2017deep}.

However, most DRL-based ASLAM approaches rely on highly simplified assumptions, including small discrete action spaces and simplified two-dimensional environments. Although these assumptions ease training, they greatly limit the applicability of such systems on real robots and scenarios.


The extremely long training time is arguably the
primary factor hindering the widespread adoption of DRL for ASLAM~\cite{placed2023survey}. In this work, we address this limitation by introducing a novel GPU-accelerated framework that enables massively parallel DRL training for Active SLAM.

By leveraging the NVIDIA Isaac Sim/Lab simulator, our approach allows hundreds of agents to be trained simultaneously (Fig.~\ref{framework}), drastically reducing training time while enabling more realistic environments and continuous action spaces. 

The main contributions of this work are:

\begin{itemize}
\item A scalable end-to-end pipeline for parallel DRL training in Active SLAM with continuous action spaces.
\item A GPU-vectorized fixed-lag SLAM backbone enabling efficient pose covariance extraction.
\item An Active SLAM formulation that uses pose covariance to guide exploration and relocalization through an uncertainty-aware reward function. 
\item A SLAM-agnostic training bridge that enables fine-tuning with arbitrary ROS2-based SLAM backbones, facilitating transfer to more realistic and complex scenarios.
\end{itemize}

\section{Related Work}

\subsection{DRL-based Active SLAM}

Placed and Castellanos~\cite{placed2020deep} present one of the first frameworks integrating model-free DRL into Active SLAM, formulating exploration as a sequential decision-making task in which the agent learns a navigation policy without explicit models of state transitions or observation probabilities. Their method uses reward functions based on true uncertainty metrics (D-optimality) and a Double Dueling Deep Q-Network (D3QN) that processes LiDAR observations and outputs discrete navigation actions. Although this work established the foundations for DRL-based ASLAM, it presents several limitations: the action space is highly discretized, training requires more than $60$ hours to converge, and experiments are restricted to simplified Gazebo environments. These limitations highlight the need for approaches that accelerate training and support more realistic scenes and complex action spaces. 

Several works have attempted to address these limitations. Alcalde \textit{et al.}~\cite{alcalde2022slam} adopt Proximal Policy Optimization (PPO) to improve training efficiency, while Zhao and Hwang~\cite{zhao2024ddpg_activeslam} employ Deep Deterministic Policy Gradient (DDPG) to introduce continuous actions, and
proposes an \textit{ad-hoc} exploration-exploitation reward, although it is not grounded in principled uncertainty metrics. Chen \textit{et al.}~\cite{chen2024lidar_end2end} propose a LiDAR-based end-to-end framework based on a dual-branch D3QN that incorporates occupancy-grid representations to improve long-term decision making. Although their method improves generalization performance in Gazebo environments and shows limited sim-to-real transfer, it still relies on a discrete action space and purely reactive policies.

Other works apply DRL only to specific components of the ASLAM pipeline rather than adopting an end-to-end formulation. In these approaches, candidate exploration locations (\textit{i.e.}, frontiers) are first generated and DRL is used to select the next target location. Chen \textit{et al.}~\cite{chen20} employ information-theoretic rewards with a DQN architecture, while Niroui \textit{et al.}~\cite{niroui19} combine optimality criteria with an Asynchronous Advantage Actor-Critic (A3C) algorithm. Similarly, Li \textit{et al.}~\cite{li19} incorporate map entropy into the reward function to guide goal selection.

\begin{table*}[t]
    \centering
    \caption{Comparison of the proposed approach and existing methods organized by the task performed by the DRL module.}
    \label{tab:comparison_prior_methods}
    \footnotesize
    \setlength{\tabcolsep}{3pt}
    \renewcommand{\arraystretch}{1.}
    \newcommand{\cmark}{\textcolor{ForestGreen}{\ding{51}}}
    \newcommand{\xmark}{\textcolor{BrickRed}{\ding{55}}}
    \begin{tabular}{l | l l l c c c c c}
        \toprule
        \textbf{Method} 
        & \makecell[l]{\textbf{Task}} 
        & \makecell[l]{\textbf{Training}\\ \textbf{Architecture}} 
        & \makecell[l]{\textbf{Reward}\\ \textbf{Design}} 
        & \makecell{\textbf{Partial}\\ \textbf{Obs.}} 
        & \makecell{\textbf{Generalization}\\ \textbf{Unknown Env.}}
        & \makecell{\textbf{Continuous} \\ \textbf{Action Space}}
        & \makecell{\textbf{Realistic}\\ \textbf{Env.} } 
        & \makecell{\textbf{Parallel} \\ \textbf{Training} }\\
        \midrule
        Niroui \textit{et al.} (2019)~\cite{niroui19} & Frontier selection & A3C & Intrinsic & \cmark & \cmark & \xmark & \cmark & \xmark \\
        Li \textit{et al.} (2019)~\cite{li19} & Frontier selection  & DQN & Map entropy & \cmark & \cmark & \xmark & \xmark & \xmark \\
        Chen \textit{et al.} (2020)~\cite{chen20} & Frontier selection & DQN  & T-opt & \cmark & \cmark & \xmark & \xmark & \xmark \\
        Cimurs \textit{et al.} (2021)~\cite{cimurs2021goal} & Local Motion & TD3 & Intrinsic & \cmark & \cmark & \cmark & \cmark & \cmark \\
        Chaplot \textit{et al.} (2020)~\cite{chaplot20} & Active localization & A3C & Posterior belief & \xmark  & \cmark & \xmark & \cmark & \xmark \\
        \midrule
        Placed \textit{et al.} (2020)~\cite{placed2020deep} & Active SLAM & D3QN & D-opt & \cmark & \cmark & \xmark & \xmark & \xmark \\
        Alcalde \textit{et al.} (2022)~\cite{alcalde2022slam} & Active SLAM & PPO & D-opt & \cmark & \cmark & \xmark  & \xmark  & \xmark 
        \\
        Zhao \textit{et al.} (2024)~\cite{zhao2024ddpg_activeslam} & Active SLAM & DDPG & Intrinsic & \cmark  & \cmark & \cmark & \xmark & \xmark          
        \\
        Chen \textit{et al.} (2024)~\cite{chen2024lidar_end2end} & Active SLAM & D3QN & Map entropy & \cmark  & \cmark & \xmark & \cmark & \xmark 
        \\ \midrule
        \textit{Ours} & Active SLAM & PPO & D-opt + Exploratory & \cmark & \cmark & \cmark & \cmark & \cmark\\
        \bottomrule
    \end{tabular}
\end{table*}

Cimurs \textit{et al.}~\cite{cimurs2021goal} propose a DRL-based exploration framework in which frontier identification and selection rely on classical information-theoretic utilities, while a Twin Delayed Deep Deterministic (TD3) policy gradient architecture is used to train the motion policy, which allows to consider a continuous action space. Although the method improves exploration efficiency, long training times limit its practical deployment. Moreover, the deterministic policy may reduce exploration capability and lead to local optima.

Due to the complexity of end-to-end ASLAM, many works adopt task decomposition strategies in which SLAM, planning, and DRL modules are trained separately. For example, Chaplot \textit{et al.}~\cite{chaplot20} train two separate policies to capture short- and long-term behaviors. Alcalde \textit{et al.}~\cite{alcalde2022slam} introduce DA-SLAM, which combines map completeness and pose uncertainty rewards but relies on discrete actions and simplified environments. Similarly, Zhao and Hwang~\cite{zhao2024ddpg_activeslam} combine Cartographer with a DDPG-based planner, while Chen \textit{et al.}~\cite{chen2024lidar_end2end} propose an end-to-end LiDAR framework based on D3QN. However, these methods typically rely on single-robot training and lack scalable parallel learning pipelines.

\subsection{Scalable Training Environments}

Training DRL policies requires a simulation environment capable of generating large amounts of interaction data. Since real-world training is impractical, simulators play a central role in the learning process. Gazebo has been widely used in robotics research due to its sensor models and integration with ROS~\cite{placed2020deep,alcalde2022slam,zhao2024ddpg_activeslam,chen2024lidar_end2end,cimurs2021goal}. However, Gazebo simulations are often simplified and relatively slow, which significantly increases DRL training time and limits scalability.

Recent GPU-accelerated simulators provide a promising alternative. NVIDIA Isaac Sim/Lab enables massively parallel simulation on GPUs, drastically accelerating training while providing realistic physics and sensor models. In a related direction, NavRL~\cite{xu2025navrl} demonstrates a large-scale training pipeline in Isaac Sim/Lab capable of training thousands of quadcopters simultaneously. The authors also introduce a bridge between Isaac Lab and ROS$2$ that allows sensor data to be exchanged through ROS topics. However, this bridge is designed primarily for inference and does not support DRL training within ROS-based pipelines.

Building on these ideas, we propose an end-to-end framework for scalable DRL training in Active SLAM. Our approach leverages GPU-based simulation to train hundreds of agents in parallel while maintaining realistic sensor models and continuous control. Additionally, we introduce a training bridge wrapper that enables fine-tuning with arbitrary ROS2-based SLAM backbones, facilitating
transfer to more realistic and complex scenarios. 

\section{Proposed Method}

\subsection{Overview}

The proposed framework addresses ASLAM as a learning control problem in which a mobile robot must simultaneously explore and map an unknown environment while maintaining a reliable localization. Unlike classical ASLAM approaches that rely on explicit information-theoretic planning at each decision step, our method learns a policy that internalizes the trade-off between exploration progress and uncertainty reduction.

\begin{figure*}[t]
      \centering
      \includegraphics[width=\textwidth]{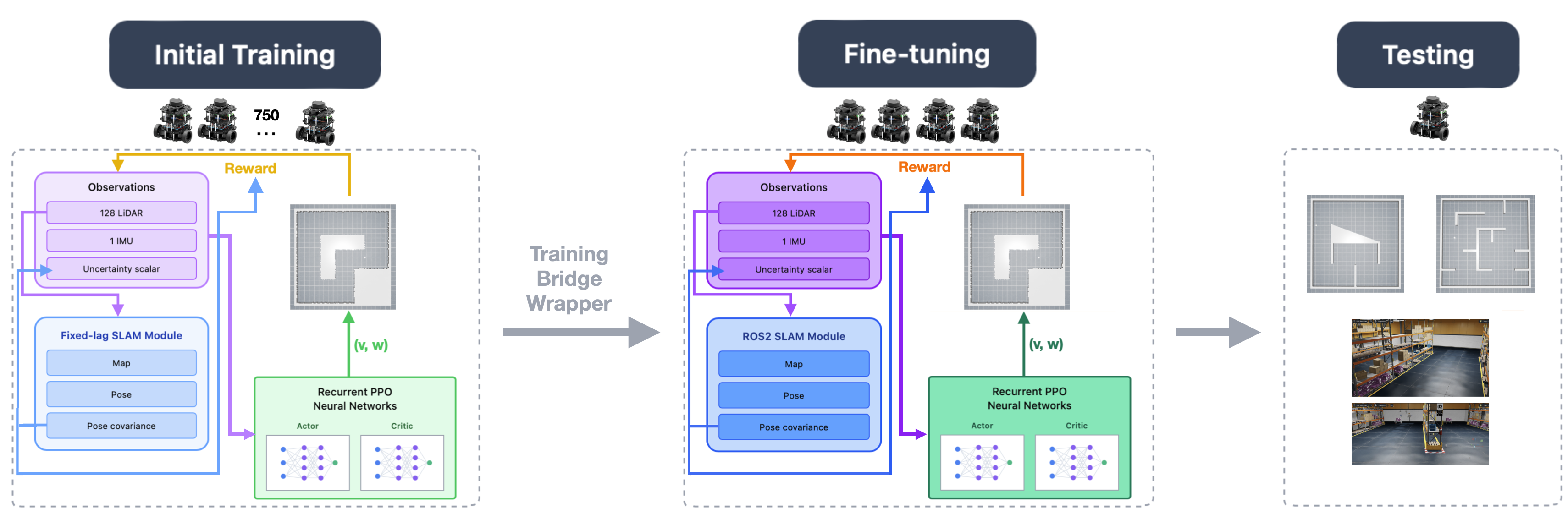}
      \caption{Overview of the proposed uncertainty-aware Active SLAM framework.}
      \label{fig:overview}
   \end{figure*}

The overall architecture is illustrated in Fig.~\ref{fig:overview}. At each timestep, raw LiDAR data are processed by a lightweight fixed-lag SLAM backbone that estimates the robot pose and its associated covariance. From this covariance matrix, we derive a compact scalar uncertainty signal that captures the confidence of the current pose estimate. This uncertainty signal is then fed to the policy network along with perceptual observations (LiDAR scans and IMU orientation).

The policy, trained using Proximal Policy Optimization (PPO) with a recurrent neural network architecture, outputs continuous linear and angular velocities. The reward function explicitly balances three objectives: (i) spatial exploration, (ii) penalization of high uncertainty regimes, and (iii) active reduction of uncertainty. This structure enables the emergence of adaptive behaviors such as relocalization when uncertainty grows and aggressive exploration when localization is reliable.
During initial training, hundreds of robots are simulated in parallel to learn an exploration policy using the GPU-accelerated fixed-lag SLAM backbone that provides a pose estimate and an uncertainty signal used in the reward function. The learned policy can then be fine-tuned with any ROS2-based SLAM system through a training bridge wrapper that progressively injects the new uncertainty signal, enabling adaptation while avoiding distribution shift. Finally, the resulting policy is evaluated in more complex and realistic testing environments.

\subsection{Problem Formulation}

Active SLAM paradigm can be formulated as a Partially Observable Markov Decision Process (POMDP):

\begin{equation}
\mathcal{P} = (S, A, T, \Omega, O, R, \gamma)\,,
\end{equation}
\noindent where:
\begin{itemize}
    \item $S$ is the latent state space, including the true robot pose $\mathcal{T}^\star\in SE(n)$ and the complete environment map~$\mathcal{M}^\star$.
    \item $A$ is the continuous action space.
    \item $T: S \times A \rightarrow \Pi(S)$ is the stochastic transition model.
    \item $\Omega$ is the observation space.
    \item $O: S \times A \rightarrow \Pi(\Omega)$ is the observation model.
    \item $R: S \times A \rightarrow \mathbb{R}$ is the reward function.
    \item $\gamma \in (0,1)$ is the discount factor.
\end{itemize}

Note that the partial observability arises because the robot never has direct access to its true pose nor to the full map; instead, it maintains a belief induced by the SLAM estimator, \textit{i.e.,}~$\mathcal{T}\in SE(n)$ and $\mathcal{M}$.

At each timestep $t$, the agent selects $a_t \in A$, receives observation $o_t \in \Omega$, and reward $R_t$ based on its new joint state. The objective is then to learn parameters $\theta$ of a stochastic policy $\pi_\theta(a|o)$ maximizing the expected discounted return:

\begin{equation}
    \pi^\star =  \arg\max_{\pi_\theta} \mathbb{E}_{\pi_\theta}
\left[
\sum_{t=0}^{\infty} \gamma^t R_t
\right] \,,
\end{equation}
\noindent where $\mathbb{E}_{\pi_\theta}$ denotes expectation over $\pi_\theta$. This formulation naturally captures the trade-off between short-term exploration gain and long-term uncertainty reduction.

\subsection{Observation and Action Spaces}
In ASLAM, the policy must reason not only about collision avoidance and navigation, but also about the evolution of localization uncertainty. Therefore, the observation space must encode both geometric structure and a compact measure of SLAM uncertainty. We use LiDAR range readings to capture local spatial geometry, the robot yaw to disambiguate orientation in partially observable settings, and a scalar uncertainty metric to explicitly inform the agent about the quality of its current state estimate. Then, the observation vector provided to the policy is:

\begin{equation}
o_t :=
\left[
\text{LiDAR}_{1:128},
\text{yaw}_t,
U_t 
\right] 
\in \mathcal{O} \,.
\end{equation}
The first 128 components correspond to range readings.
The yaw term provides orientation information.
The scalar $U_t$ encodes the SLAM pose uncertainty.

The continuous action space is defined as:

\begin{equation}
A :=
\left\{
(v, \omega) \in \mathbb{R}^2
\; | \;
v \in (0, 0.4),
\;
\omega \in (-1, 1)
\right\}\,,
\end{equation}
where $v$ and $\omega$ denote the linear and angular velocities, respectively, in m/s and rad/s.

Allowing continuous control increases realism but also makes policy optimization more challenging due to higher dimensionality and potential pathological behaviors.

\subsection{GPU-Accelerated SLAM Backbone}

Active SLAM under massively parallel DRL training imposes non-standard requirements on the estimation backend. Beyond accurate pose tracking, the system must provide a smooth and geometrically consistent uncertainty signal $U_t$ at every timestep 
and execute natively real-time on the GPU.

Full-history graph-based systems (\textit{e.g.,} g2o \cite{Kummerle2011G2O} or iSAM2-style \cite{Kaess2012iSAM2} backends) offer global consistency but incur trajectory-length–dependent computational cost, retroactive corrections, and unbounded memory growth, making them unsuitable for large-scale vectorized training. Rao–Blackwellized particle-filter methods (\textit{e.g.,} FastSLAM\cite{Montemerlo2002FastSLAM}/GMapping\cite{Grisetti2007GMapping}) expose uncertainty estimates but suffer from particle degeneracy and limited GPU efficiency due to sequential resampling. Pure scan-matching systems (\textit{e.g.,} Hector SLAM \cite{Kohlbrecher2011Hector} achieve real-time performance but lack explicit probabilistic covariance modeling. Systems such as Cartographer \cite{Hess2016Cartographer} combine scan matching with local pose-graph optimization, but introduce submaps and global loop-closure modules that break deterministic per-step computation.

To reconcile probabilistic consistency with scalable and deterministic execution, we adopt a lightweight fixed-lag pose-graph formulation that resembles the local trajectory builder of Cartographer while deliberately omitting submaps and global loop-closure constraints. This design provides constant-time updates independent of trajectory length, explicit covariance extraction from the Gauss–Newton Hessian, bounded memory through sliding-window optimization, and fully batched, vectorized GPU execution.

The environment is represented using a 2D log-odds occupancy grid, which serves both as the mapping substrate and as the reference model for correlative scan matching. This representation enables efficient probabilistic map updates, fast cell access, and GPU-friendly batched operations. Since scan-matching quality directly influences the Hessian structure,
the occupancy-grid representation shapes the statistical properties of the extracted uncertainty signal $U_t$.

We maintain a sliding-window pose graph over the last $L$ poses
\(
\mathcal{T}_{t-L:t} \subset SE(2),
\)
where each node represents the robot pose $(x,y,\theta)$ at a given timestep. Consecutive poses are connected through relative odometry factors derived from the motion model.

Each incoming LiDAR scan introduces an additional unary factor obtained via correlative scan matching. For a predicted pose $\hat{\mathcal{T}}_t$, a discretized neighborhood in $(\Delta x, \Delta y, \Delta \theta)$ is evaluated against the log-odds occupancy grid. The best-scoring candidate provides an absolute pose measurement $\tilde{\mathcal{T}}_t$, whose information matrix is scaled according to the normalized match score and the ambiguity margin between the best and second-best candidates. This weighting mitigates overconfidence under perceptual aliasing or weak geometric structure.

The resulting fixed-lag graph is optimized using a small number of Gauss–Newton iterations. Let $\mathbf{x}$ denote the stacked pose vector within the window. The local optimization minimizes:
\begin{equation}
\mathbf{x}^\star =
\arg\min_{\mathbf{x}}
\sum_{k} \| r_k(\mathbf{x}) \|_{\Omega_k}^2\,,
\end{equation}
\noindent where $r_k$ are residual terms and $\Omega_k$ are information matrices. 

The marginal covariance of the most recent pose is approximated as:
\begin{equation}
\Sigma_t \approx
\left( H_t \right)^{-1}_{\text{last block}}\,,
\end{equation}
\noindent where $H_t$ is the Gauss–Newton Hessian. This block extraction provides an efficient estimate of pose uncertainty without requiring full trajectory marginalization and directly feeds the uncertainty scalar $U_t$ introduced in the previous section.

After optimization, the aligned scan is integrated into the occupancy grid using a confidence-dependent update weight proportional to the scan-matching score. This reduces map corruption under ambiguous alignments and improves stability during training.

Finally, the entire fixed-lag SLAM pipeline is implemented in a batched and vectorized form. Multiple environments are processed simultaneously, with parallel scan-matching evaluations, batched Gauss–Newton linear system assembly, and GPU-accelerated linear algebra for Hessian inversion. This ensures that covariance extraction remains computationally negligible even when training hundreds of agents concurrently. While global loop closure is not enforced, the fixed-lag formulation provides sufficient local consistency for the uncertainty-aware Active SLAM behaviors targeted in this work.

\subsection{Reward Design}

Under the POMDP formulation of ASLAM, the agent must act to reduce belief uncertainty while expanding the coverage of the map. Since belief entropy is not optimized directly in our framework, we construct a surrogate reward that couples exploration gain with uncertainty-aware weighting. The reward function is defined as:
\begin{equation}
R_t :=
\begin{cases}
-\lambda_{col} & \text{if collision,} \\
R_{explore}^* - R_{abs} + R_{\Delta}^* & \text{otherwise,}
\end{cases}\,
\end{equation}
\noindent where $\lambda_{col} \in \mathbb{R}$ is a penalty hyperparameter chosen sufficiently large to dominate any positive reward accumulated within a short horizon, ensuring collision avoidance remains a hard constraint.
This structure enforces safety while balancing exploration and uncertainty regulation. Each term in the equation is detailed below.

\subsubsection{Exploration Reward}
\begin{equation}
R_{explore} :=
\lambda_{explore} \cdot \Delta cells \cdot I\,,
\end{equation}
where $\lambda_{explore} \in \mathbb{R}$ is an exploratory hyperparameter to balance the ASLAM reward. $\Delta cells$ measures novelty in discretized occupancy space. Newly visited cells yield higher reward than revisited ones. Specifically, the factor $\Delta cells\in[0,1]$ represents the cell reward: 1 for entering a new grid cell, and a fractional value for revisits depending on how long ago the cell was last seen, defined as:
\begin{align}
\Delta cells(x,y) := \frac{1}{2}  \frac{t(x,y)}{T_{episode}} \,,
\end{align}
where $t(x,y)$ is the number of timesteps since the last visit to the $(x,y)$ cell and $T_{episode}$ is the episode length in timesteps.
This means that revisiting a cell that was already visited during the first timesteps of the episode amounts to approximately half of its initial reward. We discretized the workspace into fixed-size cells. Also, the workspace is vastly larger than the current environment because, in SLAM, we do not know the size of our map \textit{a priori}. A cell counts as visited when the robot's ground-truth position $(x,y)$ falls inside that discretized cell. We use ground-truth information (obtained from Isaac Sim/Lab) here because if we used the SLAM pose estimate, we might count as explored cells that were not actually visited but appeared to do so due to SLAM error.

The intrinsic scaling factor $I$ is defined as:
\begin{align}
I &:= c_{t-1} + (1-c_{t-1})\cdot \iota_t \,,
\end{align}
\noindent with:
\begin{align}
c_{t-1} &:=
\text{clip}\!\left(
\frac{U_{hi} - U_{t-1}}{U_{hi}},
0,1
\right)^2 \,,\\
\iota_t &:=
\text{clip}\!\left(
\frac{\max(U_{t-1}-U_t,0)}{\kappa},
0,1
\right)\,,
\end{align}
\noindent where $\iota_t$ denotes the normalized instantaneous information gain associated with the reduction in pose uncertainty, $\kappa$ is a normalization constant that defines the uncertainty-reduction scale at which the information gain term saturates and $U_{hi}$ denotes a reference uncertainty level used to normalize the exploration gating term, corresponding to the highest uncertainty under which the exploration remains reliable.

This formulation adaptively adjusts the exploration incentive depending on the current uncertainty regime. When uncertainty is low, the coefficient $c_{t-1}$ dominates, and exploration is directly encouraged. Conversely, when uncertainty is high, the intrinsic factor increasingly depends on $\iota_t$, meaning that exploratory behavior is rewarded only if it effectively reduces uncertainty. In this way, the mechanism prevents blind exploration under poor localization while promoting informative actions that improve state estimation quality.

This implements a smooth switching mechanism between exploration and relocalization behaviors.

\subsubsection{Absolute Uncertainty Penalty}

This term penalizes operating in highly uncertain regimes.
The exponential weighting increases sensitivity in high-uncertainty states, ensuring that localization recovery becomes dominant when necessary.
\begin{equation}
R_{abs} :=
\lambda_{abs}\cdot
e^{U_{t-1}}\cdot
U_{t-1} \,,
\end{equation}
where $\lambda_{abs} \in \mathbb{R}$ is an absolute-uncertainty hyperparameter to balance the ASLAM reward.

\subsubsection{Differential Uncertainty Reward}
This term rewards reduction of uncertainty between consecutive timesteps. Positive values indicate improved localization confidence.
\begin{equation}
R_{\Delta} :=
\lambda_{\Delta}\cdot
e^{U_{t-1}}\cdot
(U_{t-1} - U_t)\, ,
\end{equation}
where $\lambda_{\Delta} \in \mathbb{R}$ is a differential-uncertainty hyperparameter to balance the ASLAM reward.

\subsubsection{Anti-Wall-Hugging Regularization}
To prevent pathological behaviors such as wall hugging, we introduce:

\begin{equation}
\alpha_{hug} :=
\text{clip}
\left(
\frac{d_{min}(t) - d_{crash}}
{d_{full} - d_{crash}},
0,1
\right)\,,
\end{equation}

where $d_{\min}(t)$ is the minimum lidar range at time t, 
$d_{\mathrm{crash}}$ is the near-collision distance threshold, and 
$d_{\mathrm{full}}$ is the safer clearance distance beyond which the wall-hugging penalty is fully removed.

Finally, the adjusted rewards become:
\begin{align}
R_{explore}^* &:= R_{explore} \cdot \alpha_{hug} \,,\\
R_{\Delta}^* &:=
\begin{cases}
R_{\Delta} \cdot \alpha_{hug}
& \text{if } U_{t-1} - U_t > 0 \\
R_{\Delta}
& \text{otherwise}
\end{cases}\,.
\end{align}

This preserves negative penalties while weakening positive incentives near obstacles.

The proposed reward formulation deliberately avoids explicitly penalizing low-speed motion or circular trajectories. Such penalties, while seemingly reasonable, often introduce unintended incentives that can lead to reward-hacking behaviors when different reward terms become inconsistent.
Instead, the structure of our reward function ensures that meaningful progress is achieved only through spatial exploration and effective uncertainty regulation. As a result, the learned policy naturally exhibits the desired behavior: it explores aggressively when localization confidence is high, and progressively shifts toward information-seeking and uncertainty-reducing actions when state estimation degrades.

This emergent balance arises from the interaction between exploration incentives and uncertainty-aware modulation, rather than from manually imposed motion constraints.

\subsection{Policy Optimization and Architecture}

\subsubsection{Optimization}

We optimize a stochastic policy $\pi_\theta$ using PPO. The policy loss is given by:
\begin{align}
r_t(\theta) &=
\frac{\pi_\theta(a_t \mid o_t)}
{\pi_{\theta_{\mathrm{old}}}(a_t \mid o_t)} \,,
\\
\bar r_t(\theta) &=
\mathrm{clip}\!\left(r_t(\theta), 1-\epsilon, 1+\epsilon\right) \,,
\\
L^{\mathrm{PPO}}(\theta)
&= - \mathbb{E}_t \left[
\min\!\left(
r_t(\theta)\hat A_t,\,
\bar r_t(\theta)\hat A_t
\right)
\right].
\end{align}
Here, $\hat A_t$ is computed via Generalized Advantage Estimation (GAE).

\subsubsection{Recurrent Layer}

We add a recurrent layer since, in order for our agent to learn Active SLAM behavior, it must have an internal representation of temporal context, in particular, whether uncertainty is growing or shrinking.

Specifically, the recurrent component is implemented as a single-layer GRU with a hidden size of $256$. We use a GRU rather than an LSTM because it provides the memory required in our pipeline while keeping the architecture lightweight.

Another way of tackling this problem would have been to increase the observation vector with additional SLAM information. However, our formulation favors generalization by exposing only a single scalar uncertainty signal $U_t$ from the SLAM backbone, while relying on the recurrent layer to capture the temporal structure necessary for decision making.

\subsection{SLAM-Agnostic Fine-Tuning via Uncertainty Bridging}

While the policy is trained using our GPU-vectorized fixed-lag SLAM backend, practical deployment may require replacing this module with an external SLAM system (\textit{e.g.,} any ROS2-based systems). Although the observation structure remains unchanged, different SLAM backbones typically produce uncertainty signals with distinct statistical distributions due to differences in modeling assumptions, scan-matching strategies, or covariance approximation schemes. Directly swapping the uncertainty source would therefore induce a distribution shift in the observation space, potentially destabilizing the learned policy.

To mitigate this effect, we introduce a training bridge wrapper that progressively injects the new uncertainty signal during fine-tuning. Let $U_t^{new}$ denote the uncertainty scalar computed by the external SLAM system. We define a blending coefficient:
\begin{equation}
\alpha_t :=
\min\left(1, \frac{t}{T_{bridge}}\right),
\end{equation}
where $T_{bridge}$ corresponds to the number of timesteps that provides a sufficiently long adaptation horizon to ensure smooth alignment between uncertainty statistics while preserving policy stability.

The uncertainty fed to the policy during fine-tuning then becomes:
\begin{equation}
U_t^{bridge} :=
\alpha_t \cdot U_t^{new} + (1-\alpha_t) \cdot U_t^{old},
\end{equation}
where $U_t^{old}$ corresponds to the uncertainty signal from the original training backend.

This gradual interpolation ensures a smooth adaptation of the policy to the new uncertainty statistics, preventing abrupt shifts in the observation distribution and stabilizing gradient updates. Importantly, this wrapper is SLAM-agnostic: any estimator providing a pose covariance can be integrated without modifying the policy architecture or reward formulation.

\section{EXPERIMENTS AND RESULTS}

\subsection{Environment Setup}

All experiments were performed on a laptop with an Intel Core i$7$ (13th Gen) CPU, an NVIDIA RTX $4060$ $8$ GB VRAM GPU, and $32$ GB RAM. We used Ubuntu $22.04$, Isaac Sim $4.5$, Isaac Lab $2.3.2$.

Similarly to \cite{placed2020deep}, a TurtleBot3 Burger robot was used, with two differential wheels and a LiDAR sensor that generates $128$ rays equally distributed over the robot’s $180^{\circ}$ front field of view, each with a minimum range of $0.1$m, a maximum range of $10$m, two decimal places of precision, and a Gaussian noise model with mean $0$ and standard deviation $0.01$. We also incorporate an IMU to our robot, in order to use its yaw information as an input.

\begin{figure}[t]
  \centering
  \captionsetup[subfloat]{labelformat=empty,font=large}
  \resizebox{0.475\textwidth}{!}{%
    \begin{minipage}{\textwidth}
      \centering
      \subfloat[Env. 2\label{fig:env2}]{%
        \includegraphics[width=0.32\textwidth]{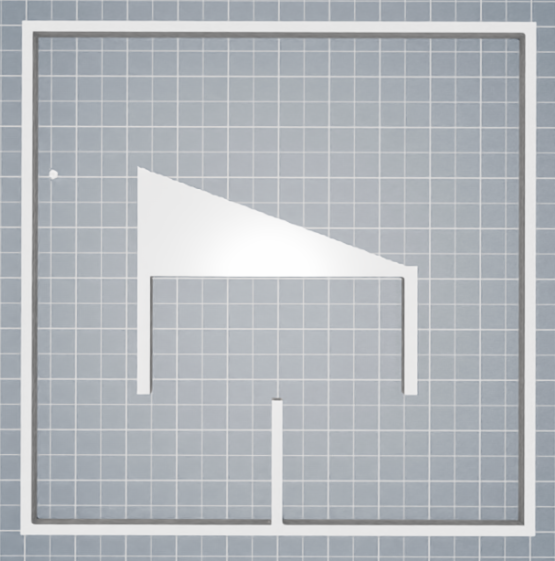}}
      \hfill
      \subfloat[Env. 3\label{fig:env3}]{%
        \includegraphics[width=0.325\textwidth]{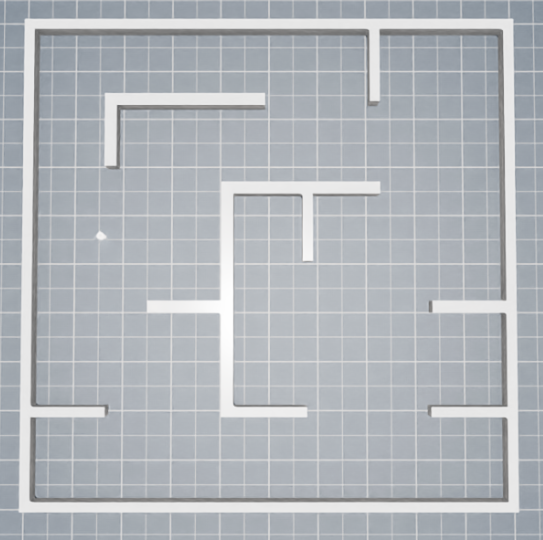}}
      \hfill
      \subfloat[Env. 4\label{fig:env4}]{%
        \includegraphics[width=0.304\textwidth]{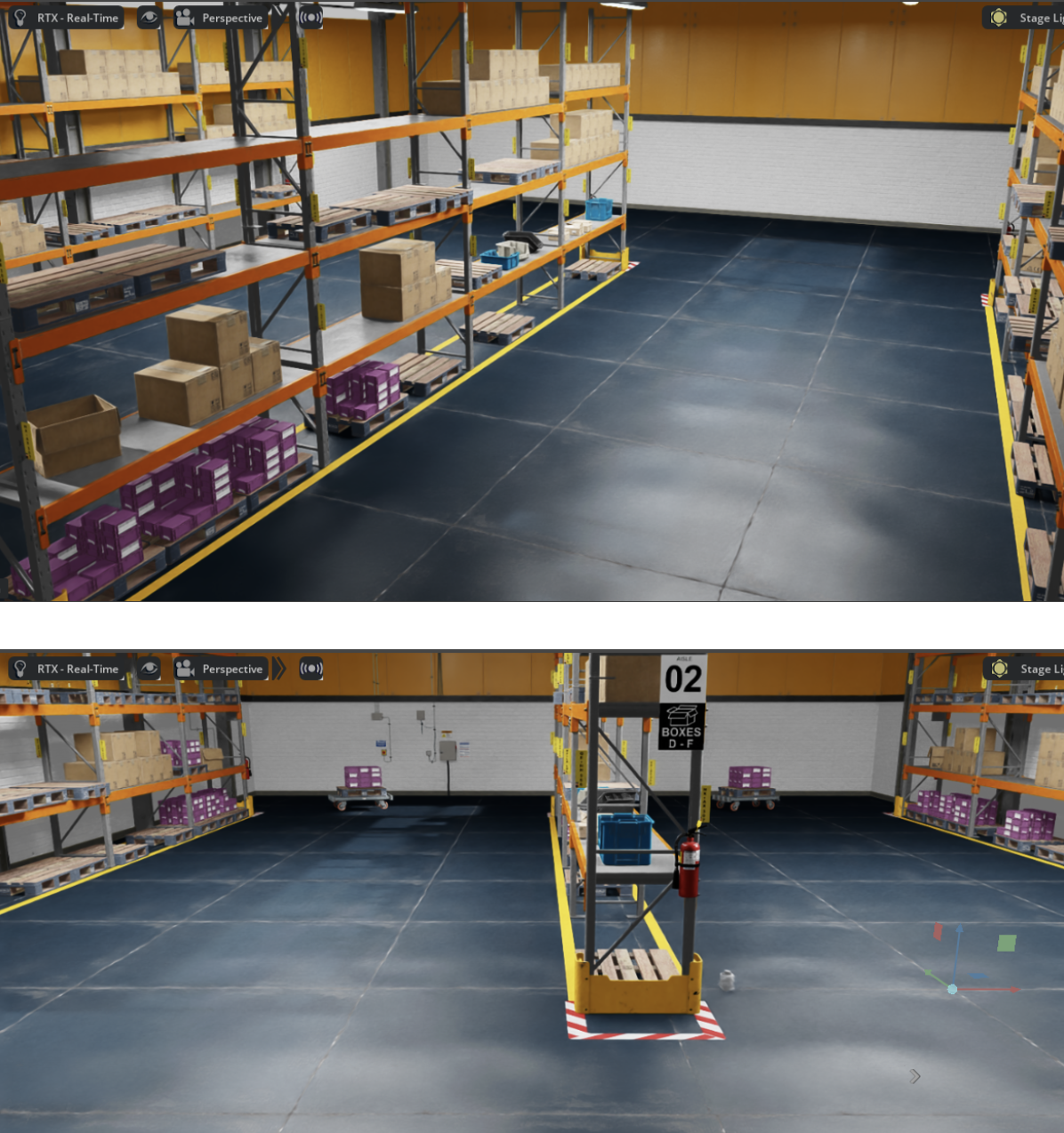}}
    \end{minipage}%
  }
  \caption{Isaac Sim/Lab testing environments. Env.~2 and Env.~3 were recreated following~\cite{placed2020deep}. Env.~4 was taken from~\cite{nvidia_isaacsim_env_assets_4_5_0}.}
  \label{fig:envs}
\end{figure}

Following \cite{placed2020deep} and \cite{alcalde2022slam}, which use three different environments created in Gazebo, we recreated them on Isaac Sim/Lab. The first environment (see Fig. \ref{framework}), where our agent is trained, is a simple layout with repeated turns. The second and third environments, which are more difficult (see Env. 2 and Env. 3 in Fig. \ref{fig:envs}), were used only for testing. We also increased the size of the environments relative to the TurtleBot$3$ compared to \cite{placed2020deep} to allow more maneuverability. We made the walls of the first environment ``rougher" to reduce perceptual aliasing, as our SLAM is LiDAR-only. This trick, while cheap, improved the training results without hindering computational performance. Since these environments, although useful, are far from real-life environments, we included \cite{nvidia_isaacsim_env_assets_4_5_0}, a built-in realistic warehouse, as a testing environment, exploiting the capabilities of the simulator Isaac Sim (see Env. 4 in Fig.~\ref{fig:envs}).

The training of our agent consisted of episodes in which the robot navigates until $2048$ timesteps are used or a collision occurs (i.e. gets closer to any obstacle than a defined threshold of $0.15$m). Parallel training was performed using $750$ agents simultaneously. The training and reward hyperparameter settings can be found on our project’s GitHub.

\begin{figure}[t]
  \centering
  \includegraphics[width=0.49\columnwidth]{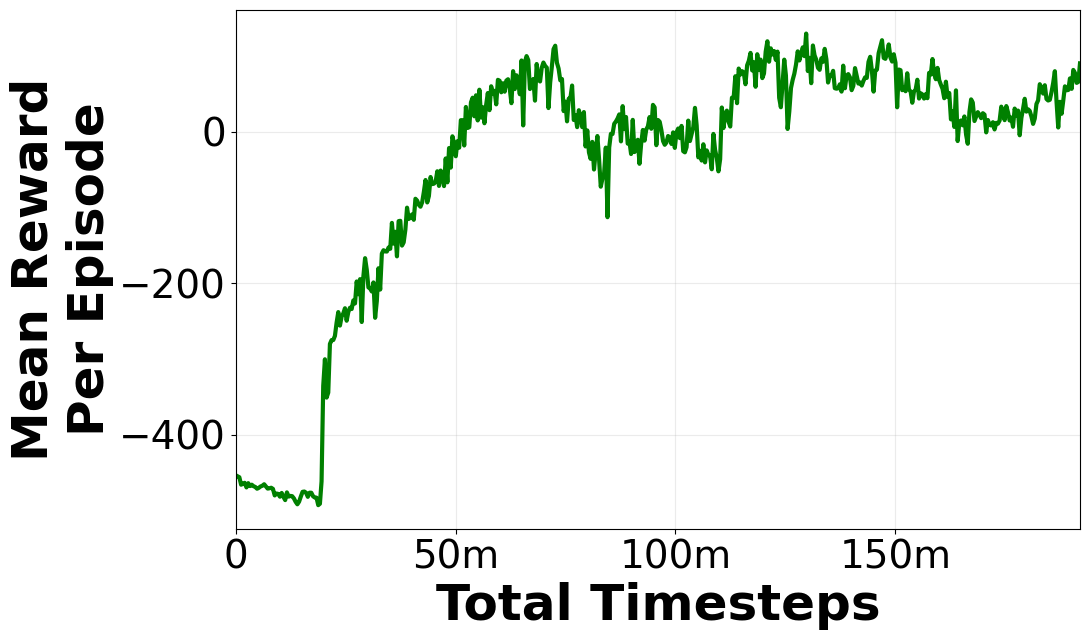}\hfill
  \includegraphics[width=0.49\columnwidth]{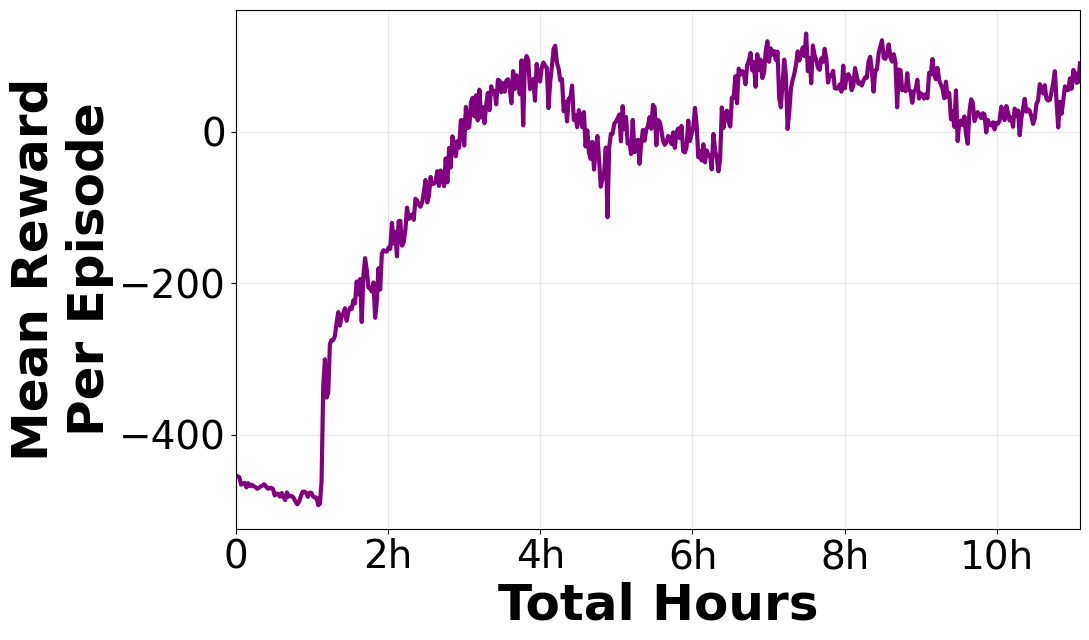}
  \caption{Training curves of our pipeline using $750$ parallel agents in environment \ref{framework}. This figure shows that tens of millions of ASLAM training timesteps can be achieved in a few hours with our pipeline.}
  \label{fig:train}
\end{figure}

\begin{figure}[t]
  \centering
  \includegraphics[width=0.6\columnwidth]{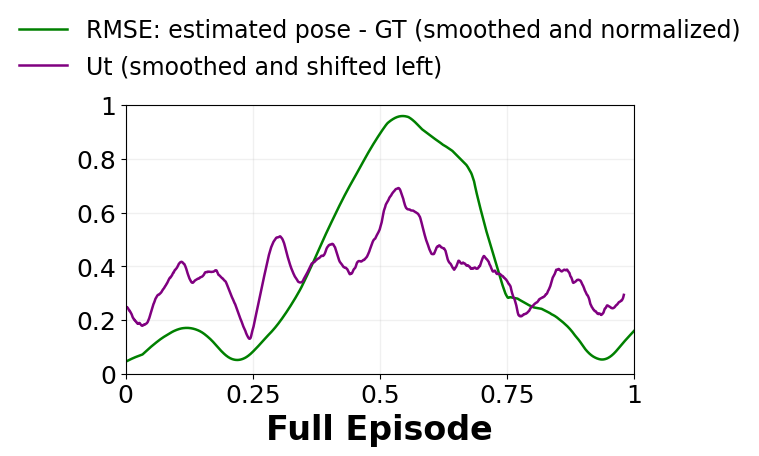}
  \caption{Correlation between RMSE (estimated pose - GT) and $U_t$. This graph was computed in Environment $1$.}
  \label{fig:RMSE-Ut}
\end{figure}

\begin{figure}[t]
  \centering
  \resizebox{0.475\textwidth}{!}{
  \subfloat[Env. 1\label{fig:left}]{%
    \includegraphics[width=0.3175\textwidth]{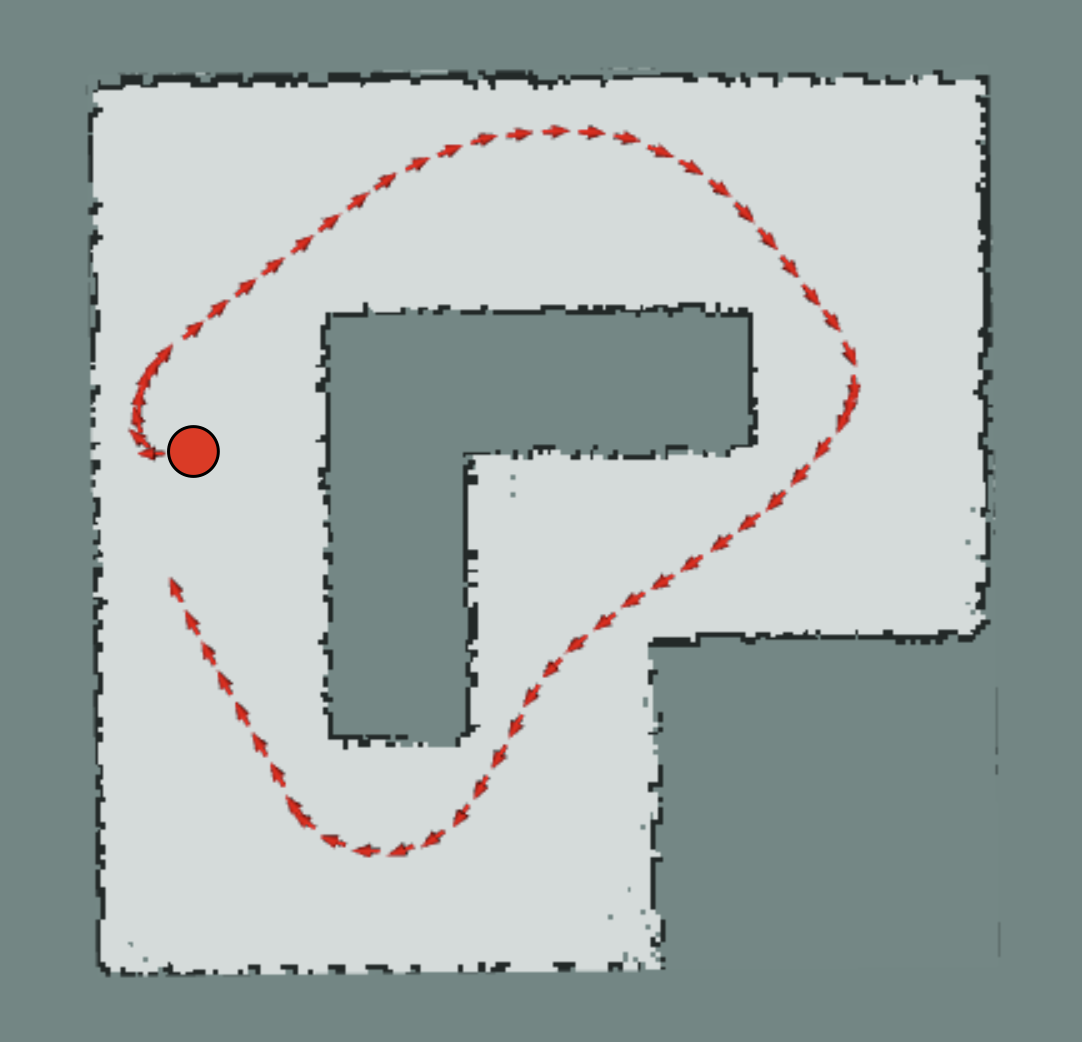}}
  \hfill
  \subfloat[Env. 2\label{fig:center}]{%
    \includegraphics[width=0.3175\textwidth]{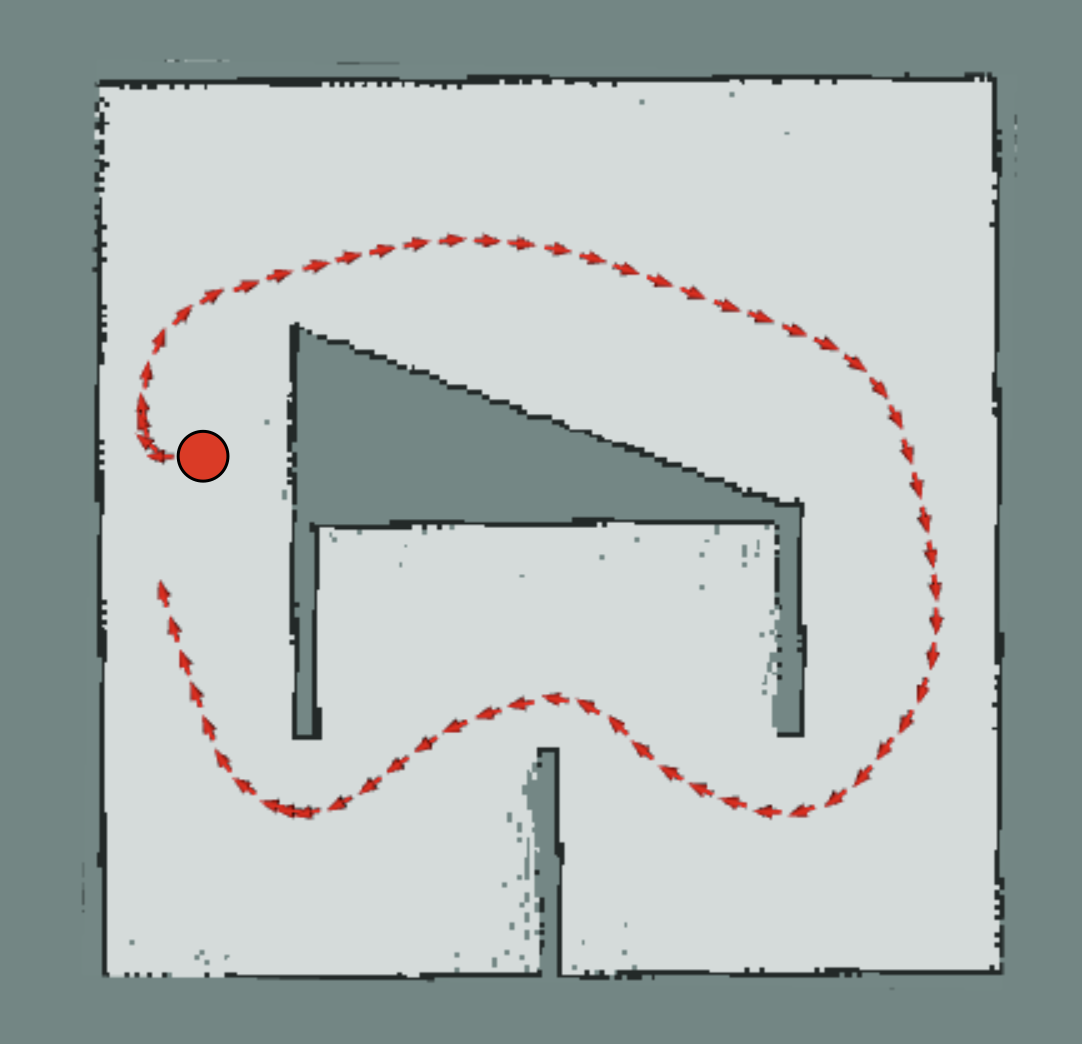}}
  \hfill
  \subfloat[Env. 3\label{fig:right}]{%
    \includegraphics[width=0.32525\textwidth]{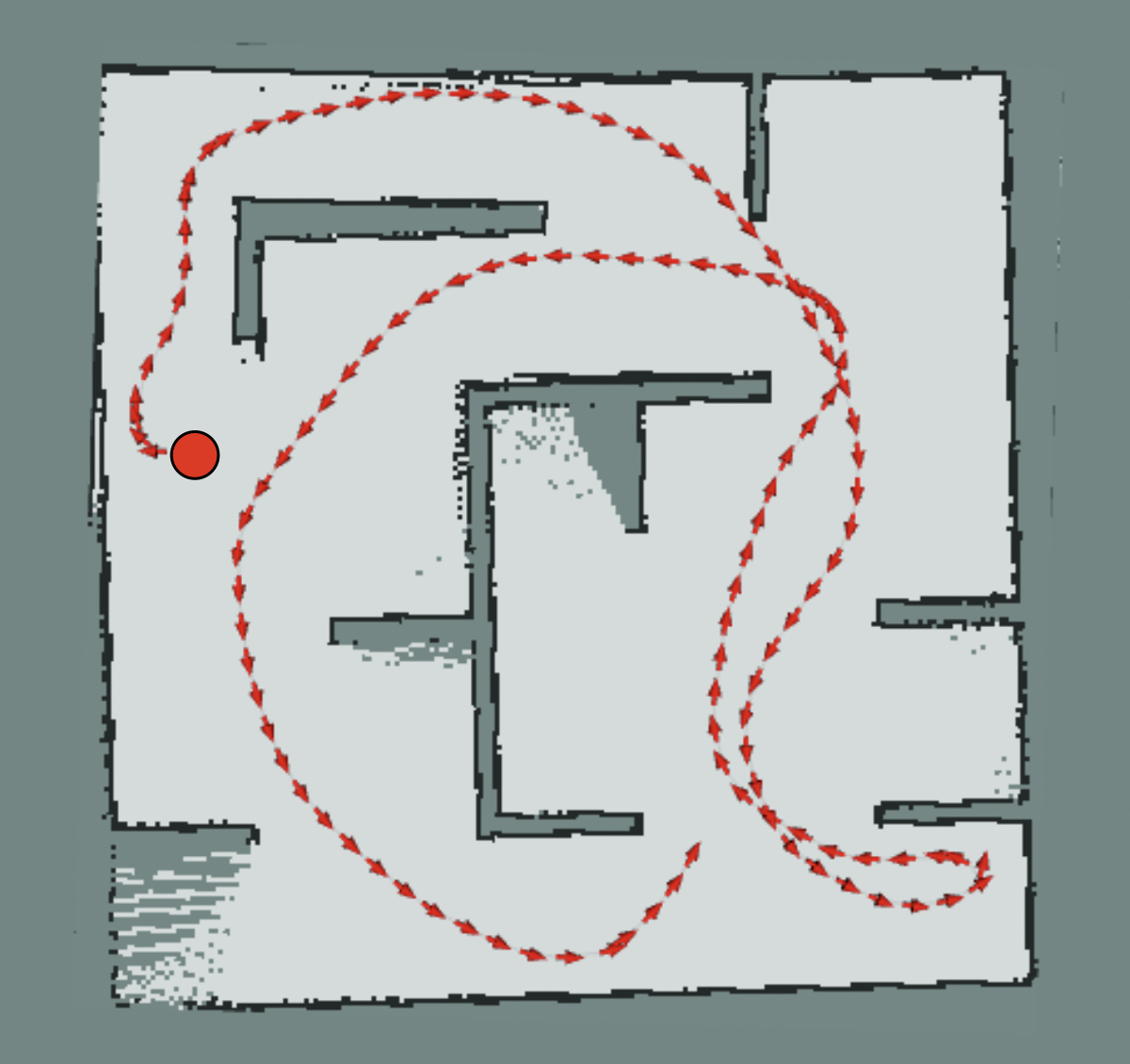}}
    }
  \caption{Occupancy grids created by our fine-tuned uncertainty-aware agent using \texttt{slam\_toolbox} and \texttt{rviz2}. Red circles indicate the starting position, whereas red arrows indicate the trajectory followed by the robot.}
  \label{ogs}
\end{figure}


\subsection{Discussion of Results}

We can visualize the respective training curves in Fig. \ref{fig:train}, which confirms the feasibility of using DRL for Active SLAM in a reasonable time. In \cite{placed2020deep} and \cite{alcalde2022slam}, more than $50$ hours were required for a simpler policy to emerge; in contrast, the policy we used for our experiments took approximately $4$ hours to train on our average laptop.

In Fig.~\ref{fig:RMSE-Ut}, we can visualize two curves:
\begin{itemize}
\item the smoothed and normalized Root Mean Squared Error (RMSE) between the estimated pose from our fixed-lag SLAM and the ground truth from the simulator;
\item our uncertainty signal $U_t$ (smoothed and shifted to the left).
\end{itemize}
Although $U_t$ is noisier, the correlation between the two curves is clear. In fact, $$corr(RMSE, U_{t+shift}) = 0.771956 \,$$ where $corr$ represents the Pearson correlation coefficient. A non-trivial comment is that $U_t$ follows the structure of the RMSE but with a delay. This can be interpreted as ``first the robot starts getting lost; then it becomes uncertain about its localization,'' which intuitively makes sense. Because of this inherent delay, the recurrent layer becomes a fundamental piece of our pipeline, allowing our agent to capture this temporal context.

In Table \ref{tab:env-results}, we can see a comparison between:
\begin{itemize}
    \item PPO$_{exploratory}$, a trained agent that completely neglects covariance both as an observation and as a reward;
    \item PPO$_{uncertainty}$, which is the full version of our agent and considers exploratory, uncertainty, and anti-wall-hugging rewards.
\end{itemize}
We observe that our PPO$_{uncertainty}$ agent has a significantly lower collision rate, and that its best run explored a larger percentage of the entire environment, where we define this exploration metric as the percentage of the environment “seen” by the robot’s LiDAR.

We also compared these two trained agents with a random agent and a classical frontier-based, non-learning exploration agent, as in~\cite{yamauchi1997frontier}, but these results were not included because both performed extremely poorly in our experiments.

Finally, we performed a fine-tuning of our PPO$_{uncertainty}$ agent using our training bridge wrapper along with \texttt{slam\_toolbox} \cite{slam_toolbox}, a full SLAM system that includes a global loop-closure module.
In Fig.~\ref{ogs}, we can visualize the occupancy grids along with the trajectories created by our fine-tuned uncertainty-aware agent using the \texttt{slam\_toolbox} and \texttt{rviz2} \cite{rviz2} ROS $2$ packages. The trajectories show that the agent tends to follow structured exploration patterns that progressively expand the explored region while maintaining localization stability. In the most complex environment (Env. 3) the robot revisits previously explored areas, implicitly performing loop-like motions that help reduce pose uncertainty and improve map consistency. This behavior results from the uncertainty-aware ASLAM formulation. 

\begin{table}[t]
  \caption{Test results across three environments letting the agent move for a maximum of $3000$ timesteps ($5$ minutes) or until a collision occurs. The results were averaged across $100$ runs. The collision rate represents actual physical collisions against obstacles.}
  \label{tab:env-results}
  \centering
  \resizebox{0.99\columnwidth}{!}{
  \begin{tabular}{c c c c c}
    \toprule
    \textbf{Environment} & \textbf{Agent} &
    \textbf{Unique visited cells} &
    \textbf{Best run map explored (\%)} &
    \textbf{Collision rate} \\
    \midrule
    \multirow{2}{*}{\makebox[0.12\textwidth][c]{\includegraphics[width=0.09\linewidth]{env2.png}}} &
      PPO$_{exploratory}$  & 412 & 100 & 0 \\
    \cmidrule(lr){2-5}
    & PPO$_{uncertainty}$ & 464 & 100 & 0 \\
    \midrule
    \multirow{2}{*}{\makebox[0.12\textwidth][c]{\includegraphics[width=0.09\linewidth]{env3.png}}} &
      PPO$_{exploratory}$  & 482 & 87 & 24\% \\
    \cmidrule(lr){2-5}
    & PPO$_{uncertainty}$ & 416 & 92 & 11\% \\
    \midrule
    \multirow{2}{*}{\makebox[0.12\textwidth][c]{\includegraphics[width=0.085\linewidth]{env4.png}}} &
      PPO$_{exploratory}$  & 736 & 100 & 3\% \\
    \cmidrule(lr){2-5}
    & PPO$_{uncertainty}$ & 717 & 100 & 1\% \\
    \bottomrule
  \end{tabular}%
  }
\end{table}

\section{CONCLUSIONS}

In this work, we present the first framework for Massive Parallel Deep Reinforcement Learning applied to Active SLAM. Our main contribution is a complete end-to-end system enabling scalable and massively parallel training in the Isaac Sim/Lab simulator. In addition, we included a training bridge wrapper, which allows the fine-tuning of our already trained agents with any ROS $2$ SLAM package. 

We showed that our method not only accelerates training but also allows for the use of a continuous action space and the exploration of more challenging scenarios. This approach drastically reduces training time, achieving the desired behavior in approximately $4$ hours, compared to the more than $50$ hours estimated in prior works.

The training bridge wrapper is optimized for live data transfer between Isaac Lab and ROS $2$ during training and is flexible enough to be used in other robotic tasks that require the use of DRL, Isaac Lab and ROS $2$ in sync. 

The developed framework has been publicly released to ensure experiment repeatability and foster community adoption. We consider this work a key milestone toward making autonomous exploration and Active SLAM with DRL scalable and efficient.

\bibliographystyle{IEEEtran}
\bibliography{IEEEabrv,biblio}
\newpage

\end{document}